\documentclass[letterpaper, 10 pt, conference]{IEEEconf}
\IEEEoverridecommandlockouts
\usepackage{cite}
\usepackage{amsmath,amssymb,amsfonts}
\usepackage{algorithmic}
\usepackage{graphicx}
\usepackage{textcomp}
\usepackage{xcolor}
\usepackage[left=1.91cm, right = 1.91cm, top=1.91cm, bottom=1.93cm]{geometry}
\usepackage{hyperref}
\let\labelindent\relax
\usepackage{enumitem}
\usepackage[font=normalsize,labelfont=bf]{caption}
\def\BibTeX{{\rm B\kern-.05em{\sc i\kern-.025em b}\kern-.08em
    T\kern-.1667em\lower.7ex\hbox{E}\kern-.125emX}}

\begin{document}
\title{\LARGE \bf 
A Co-Design Framework for Energy-Aware Monoped Jumping with Detailed Actuator Modeling\\}

\author{Aman Singh$^{*,1}$, Aastha Mishra$^{*,1}$, Deepak Kapa$^{2}$,  Suryank Joshi$^{3}$, and Shishir Kolathaya$^{4}$% <-this % stops a space
\thanks{* These authors have contributed equally. This work is supported by AI and Robotics Technology Park (ARTPARK).}
\thanks{$^{1}$ Aman Singh and Aastha Mishra are with the Robert Bosch Center for Cyber Physical Systems, Indian Institute of Science, Bengaluru.}
\thanks{$^{2}$ Deepak Kapa is with Indian Institute of Technology, Roorkee.}
\thanks{$^{3}$ Suryank Joshi is with Manipal Institute of Technology, Manipal.}
\thanks{$^{4}$Shishir Kolathaya is with the Robert Bosch Center for Cyber Physical Systems and the Department of Computer Science \& Automation, Indian Institute of Science, Bengaluru. 
{\tt\small stochlab@iisc.ac.in}
\thanks{This work was supported by the AI \& Robotics Technology Park (ARTPARK) at IISc.}% <-this % stops a space
% (email: \href{mailto:stochlab@iisc.ac.in}{stochlab@iisc.ac.in})
}
}

% \title{Conference Paper Title*\\
% {\footnotesize \textsuperscript{*}Note: Sub-titles are not captured in Xplore and
% should not be used}
% \thanks{Identify applicable funding agency here. If none, delete this.}
% }

% \author{\IEEEauthorblockN{1\textsuperscript{st} Given Name Surname}
% \IEEEauthorblockA{\textit{dept. name of organization (of Aff.)} \\
% \textit{name of organization (of Aff.)}\\
% City, Country \\
% email address or ORCID}

% \and

% \IEEEauthorblockN{2\textsuperscript{nd} Given Name Surname}
% \IEEEauthorblockA{\textit{dept. name of organization (of Aff.)} \\
% \textit{name of organization (of Aff.)}\\
% City, Country \\
% email address or ORCID}

% \and

% \IEEEauthorblockN{3\textsuperscript{rd} Given Name Surname}
% \IEEEauthorblockA{\textit{dept. name of organization (of Aff.)} \\
% \textit{name of organization (of Aff.)}\\
% City, Country \\
% email address or ORCID}

% \and

% \IEEEauthorblockN{4\textsuperscript{th} Given Name Surname}
% \IEEEauthorblockA{\textit{dept. name of organization (of Aff.)} \\
% \textit{name of organization (of Aff.)}\\
% City, Country \\
% email address or ORCID}

% \and

% \IEEEauthorblockN{5\textsuperscript{th} Given Name Surname}
% \IEEEauthorblockA{\textit{dept. name of organization (of Aff.)} \\
% \textit{name of organization (of Aff.)}\\
% City, Country \\
% email address or ORCID}

% \and

% \IEEEauthorblockN{6\textsuperscript{th} Given Name Surname}
% \IEEEauthorblockA{\textit{dept. name of organization (of Aff.)} \\
% \textit{name of organization (of Aff.)}\\
% City, Country \\
% email address or ORCID}

\maketitle
\thispagestyle{empty}
\pagestyle{empty}

\begin{abstract} 
A monoped’s jump height and energy consumption depend on both, its mechanical design and control strategy. Existing co-design frameworks typically optimize for either maximum height or minimum energy, neglecting their trade-off. They also often omit gearbox parameter optimization and use oversimplified actuator mass models, producing designs difficult to replicate in practice.
In this work, we introduce a novel three-stage co-design optimization framework that jointly maximizes jump height while minimizing mechanical energy consumption of a monoped. The proposed method explicitly incorporates realistic actuator mass models and optimizes mechanical design (including gearbox) and control parameters within a unified framework. The resulting design outputs are then used to automatically generate a parameterized CAD model suitable for direct fabrication, significantly reducing manual design iterations. Our experimental evaluations demonstrate a 50\% reduction in mechanical energy consumption compared to the baseline design, while achieving a jump height of 0.8m. 
Video presentation is available at: 
\href{http://y2u.be/XW8IFRCcPgM}{http://y2u.be/XW8IFRCcPgM}
\end{abstract}

% \begin{IEEEkeywords}

% \end{IEEEkeywords}

% \vspace{-1em}
\section{Introduction}

Legged robots are pivotal for navigating complex, unstructured terrains where wheeled or tracked systems fail. 
% Inspired by animal locomotion \cite{DesignPrinciples}, they offer superior mobility and have potential for disaster response and planetary exploration. 
Jumping enables them, including monopeds, to overcome obstacles or terrain gaps. Several quadrupeds \cite{MitCheetah3, MiniCheetah, StanfordDoggo} and humanoids \cite{BostonDynamicsAtlas, UnitreeHumanoid} demonstrate high jumps, while monopeds provide a simpler, highly dynamic platform for performance optimization. Power efficiency remains crucial, as minimizing energy use during jumps is essential for deployment.

% While prior studies have explored robotic jumping, most rely on control strategies, with mechanical design chosen heuristically. However, optimal jumping performance requires simultaneous optimization of both mechanical structure and control. For instance, \cite{PantherLeg} optimizes actuator design for jump height but neglects energy consumption and does not consider link lengths as variables. Recent co-design approaches \cite{StarlETHCooptimization, Vitruvio, MetaRLCodesign} incorporate mechanical and control optimization, addressing parameters like link lengths, transmission ratios, and spring stiffness, though they focus on walking rather than jumping. Similarly, \cite{Co-designing_versatile_quadruped_robots_for_dynamic_and_energy-efficient_motions} emphasizes energy efficiency at a fixed jump height, limiting performance. Work such as \cite{Computational_Design___Morphological_Computation} maximizes jump height using passive components like springs to improve energy efficiency, but without optimizing control parameters.

Most prior studies on robotic jumping focus on control strategies with heuristic mechanical design, whereas optimal performance requires joint optimization of mechanical structure and control. For example, \cite{PantherLeg} optimizes actuator design for jump height but ignores energy consumption and link length variation. Recent co-design methods \cite{StarlETHCooptimization, Vitruvio, MetaRLCodesign} optimize link lengths, transmission ratios, and spring stiffness, but target walking rather than jumping. Similarly, \cite{Co-designing_versatile_quadruped_robots_for_dynamic_and_energy-efficient_motions} improves energy efficiency at a fixed jump height, limiting performance. Work such as \cite{Computational_Design___Morphological_Computation} maximizes jump height using passive elements like springs to improve efficiency, yet omits control parameter optimization.

\begin{figure}
    \centering    
    \includegraphics[width=0.9\linewidth]{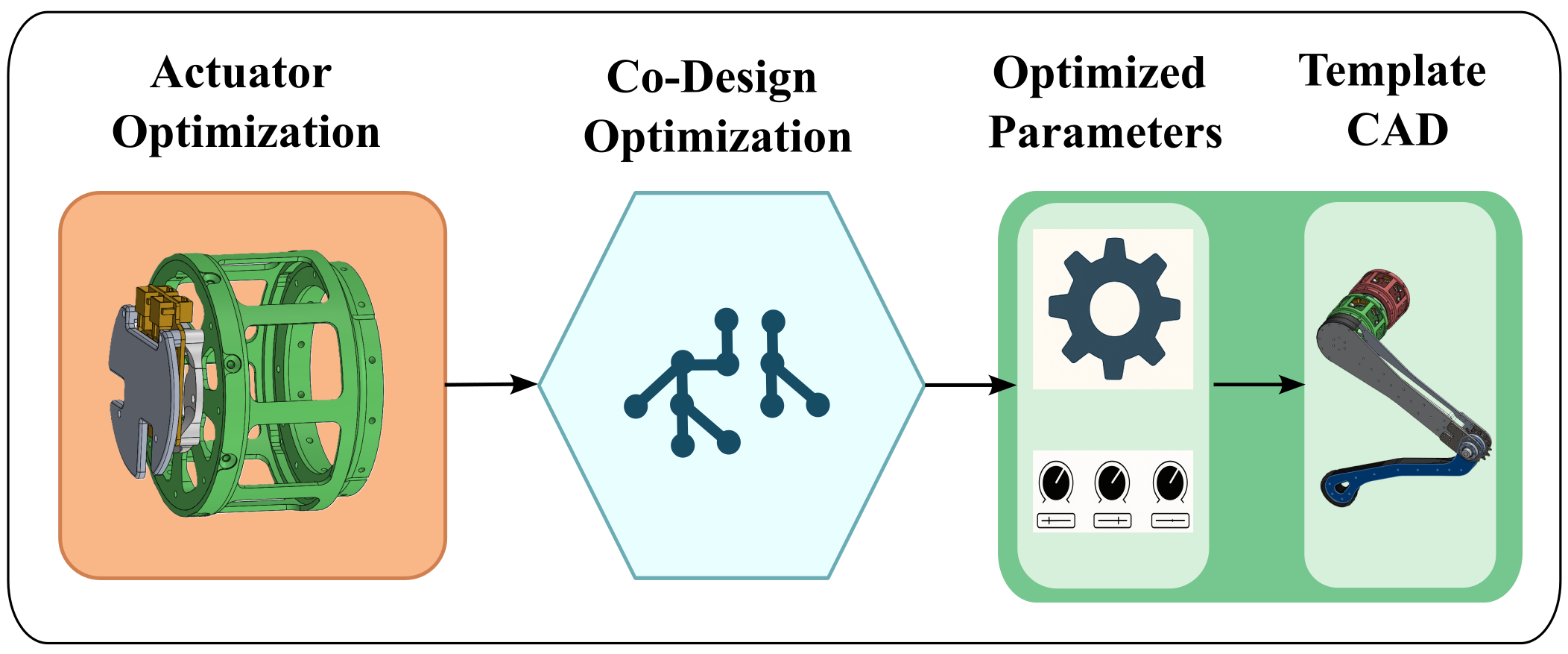}
    \caption{Overview of the proposed methodology}
    \label{fig:Intro_Diagram}
    \vspace{-1.5em}
\end{figure}

% While prior work has explored co-design for legged systems, many studies have given limited attention to actuator gearbox modeling, particularly in the context of optimizing both jumping performance and energy consumption. Studies such as \cite{MetaRLCodesign, Vitruvio, StarlETHCooptimization, Co-designing_versatile_quadruped_robots_for_dynamic_and_energy-efficient_motions} primarily focus on link lengths, transmission ratios, and spring stiffness, without considering the detailed design of planetary gearboxes. Similarly, \cite{JointOptIFT} addresses optimization of link lengths and actuator attachment points but omits gearbox design entirely. Co-design efforts applied to manipulators and monopeds in \cite{StochasticProgCodesign2, StochasticProgCodesign1} consider gear ratios, compliance, and link masses, yet do not evaluate gearbox type, internal parameters, or actuator mass. Other studies such as \cite{A_versatile_co-design___legged_robots, Computational_design_of____size_and_actuators, Simulation_Aided_Co-Design_for_Robust_Robot_Optimization} include models for motor and gearbox friction but are tailored to belt-driven systems rather than planetary configurations. Although actuator gearbox parameters are optimized in works like \cite{PantherLeg, KaistHound}, they do not account for the overall actuator mass or include joint control parameter optimization.
% which are essential for energy-aware performance-driven design.

While co-design for legged systems has been studied, actuator gearbox modeling, especially for jointly optimizing jumping performance and energy consumption, remains underexplored. Works such as \cite{MetaRLCodesign, Vitruvio, StarlETHCooptimization, Co-designing_versatile_quadruped_robots_for_dynamic_and_energy-efficient_motions} focus on link lengths, transmission ratios, and spring stiffness, but omit detailed planetary gearbox design. Similarly, \cite{JointOptIFT} optimizes link lengths and actuator attachment points without addressing gearboxes. Co-design efforts for manipulators and monopeds \cite{StochasticProgCodesign2, StochasticProgCodesign1} consider gear ratios, compliance, and link masses, yet exclude gearbox type, internal parameters, or actuator mass. Other studies \cite{A_versatile_co-design___legged_robots, Computational_design_of____size_and_actuators, Simulation_Aided_Co-Design_for_Robust_Robot_Optimization} model motor and gearbox friction but target belt-driven systems which are non-planetary. Although \cite{PantherLeg, KaistHound} optimize gearbox parameters, they omit overall actuator mass and joint control optimization.

\begin{figure*}[ht]
    \centering    
    \includegraphics[width=\linewidth]{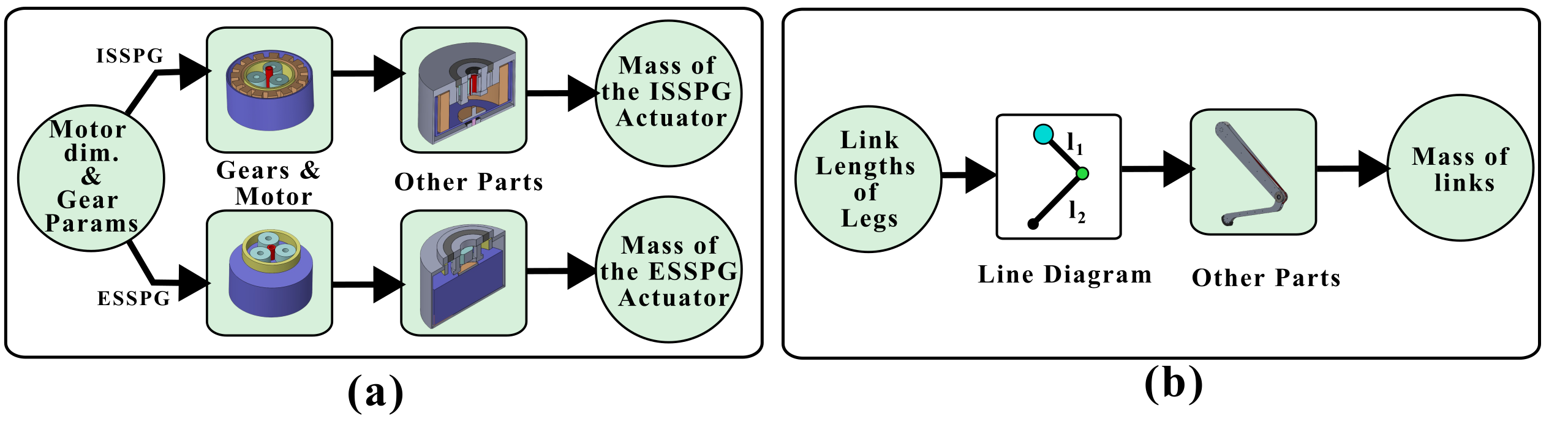}
    \vspace{-1em}
    % \caption{Mass modeling of actuators and leg links. (a) Actuator mass model for ISSPG and ESSPG types: given motor dimensions and gear parameters, the dimensions of all actuator components are computed to estimate total mass. (b) Leg-link mass model: given the link lengths \(l_1\) and \(l_2\), dimensions of the remaining parts are derived to compute the total mass of the leg links.}
    \caption{Mass modeling of actuators and leg links. (a) Actuator mass model for ISSPG and ESSPG: motor dimensions and gear parameters determine all component dimensions, yielding total mass. (b) Leg-link mass model: link lengths $l_1$ and $l_2$ define remaining part dimensions to compute total leg-link mass.}
    \label{fig:Mass_models}
    \vspace{-1em}
\end{figure*}

%%%%%%%%%%%%%%%%%%%%
%%%%% Research Gap
%%%%%%%%%%%%%%%%%%%%

% From the reviewed literature, it is evident that although several studies address the simultaneous optimization of design and control parameters, they often focus exclusively on either performance or energy efficiency, rather than jointly optimizing both objectives. Furthermore, current co-optimization approaches typically overlook the inclusion of detailed gearbox parameters and gearbox type as design variables.
% % despite their critical role in the effective design of monopeds and other legged robots. 
% Additionally, existing methods rely on nominal models for calculating the mass of components such as motors, gearboxes, and links, resulting in significant discrepancies between estimated and actual masses.
% % thereby limiting the accuracy and applicability of the computational design frameworks.
% To address the identified research gap, we develop a co-design optimization framework that concurrently determines optimal design and control parameters for a monoped robot. To summarize, the key contributions of this work are:
% %\vspace{-0.1em}

The literature shows that many studies optimize design and control parameters, but often target either performance or energy efficiency, not both. Most co-optimization approaches omit detailed gearbox parameters and gearbox type as design variables. Existing methods also rely on nominal models to estimate component masses (motors, gearboxes, and links) leading to notable deviations from actual values. To bridge this gap, we present a co-design optimization framework for a monoped robot. The key contributions of this work are:

\begin{itemize}
    \item We propose a novel co-design optimization framework that maximizes height in a jump while minimizing energy consumption by jointly optimizing design and control parameters.
    
    % \item We incorporate detailed gearbox parameters, including number of teeth, module, number of planet gears, and actuator type—Internal Single Stage Planetary Gearbox (ISSPG) and External Single Stage Planetary Gearbox (ESSPG). To the best of our knowledge, this is the first instance of optimizing detailed gearbox parameters and gearbox type within a co-design optimization framework.

    \item We include gearbox parameters in detail like, number of teeth, module, number of planet gears, and actuator type (Internal and External configuration). To our knowledge, this is the first optimization of gearbox parameters and type within a co-design framework.
    
    \item We utilize realistic mass models for hip and knee actuators that incorporate gears, motors, and components like bearings, carriers, casings, and backplates. Link mass models include all constituent parts, offering improved accuracy over previous methods. 
    % \color{red}{This makes our framework more amenable to deployment in real-world systems.}
\end{itemize}

% The structure of the paper is as follows: Section \ref{leg_design} describes the design and control architecture of the monoped, whose control and design parameters are further optimized using the proposed framework. Section \ref{opt_framework} outlines the methodology, including the proposed co-design framework and the formulation of the optimization objectives. Section \ref{results} presents the optimization results. Finally, section \ref{conclusion} summarizes the findings and discusses potential directions for future research.

The paper is organized as follows: Section \ref{leg_design} describes the monoped’s design and control architecture. Section \ref{opt_framework} outlines the methodology, including the co-design framework and optimization objectives. Section \ref{results} presents the results, and Section \ref{conclusion} summarizes the findings and future research directions.
%\vspace{-0.5em}
\vspace{-1em}
\section{Preliminaries}\label{leg_design}
This section details the monoped’s design and leg-link mass models, and the control architecture for jumping.

\begin{figure}[ht]
    \centering
    \includegraphics[width=\linewidth]{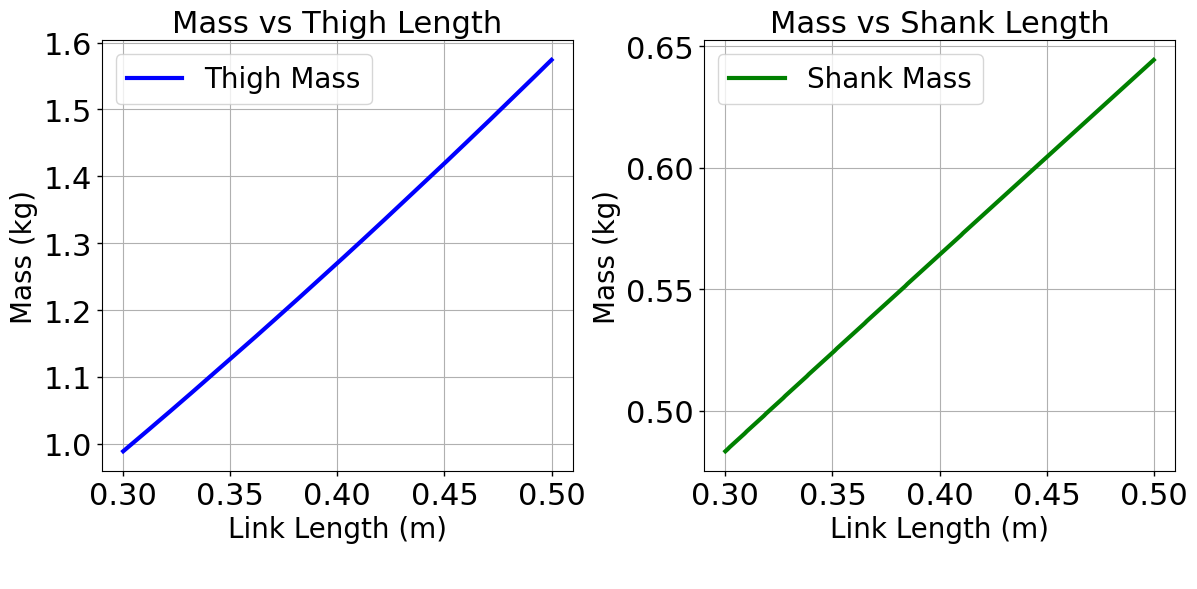}
    \vspace{-1em}
    \caption{Link-length to mass mapping, derived from the leg link mass model described in Section~\ref{monoped arch}.}
    \label{fig:mass_v_ll}
    \vspace{-1em}
\end{figure}

\subsection{Monoped Design Architecture}
\label{monoped arch}
The monoped uses a $2$-DOF leg with actuated hip and knee joints, both placed proximally near the hip to reduce distal mass. A chain-driven transmission actuates the knee, lowering leg inertia and improving jumping performance.

High gear ratio actuators increase torque density and reduce motor load but raise reflected inertia and damping \cite{DesignPrinciples}, limiting performance in dynamic tasks like jumping. Direct-drive actuators offer low reflected inertia but lack sufficient torque and efficiency in high-torque, low-speed regimes due to Joule heating \cite{design_principles_direct_drive}. Quasi-Direct Drive (QDD) actuators \cite{MitCheetah3,MiniCheetah,CheetahActuator} balance these trade-offs by pairing a low gear ratio with the motor, providing adequate torque while keeping reflected inertia low for dynamic motion. 

Accordingly, each monoped actuator uses a QDD setup with a brushless DC (BLDC) motor and single-stage planetary gearbox. Multiple stages are avoided to prevent increased inertia and reduced efficiency\cite{Design_Principles_TMech}, and alternative drives (Harmonic, Cycloidal) are excluded as they suit high gear ratios. Two planetary gearbox types are used: External Single-Stage Planetary Gearbox (ESSPG)~\cite{MitCheetah3}, mounted outside the motor, and Internal Single-Stage Planetary Gearbox (ISSPG)~\cite{MiniCheetah}, integrated within the stator. ISSPG is lighter and more compact but limited in ratio, while ESSPG is heavier but supports higher ratios~\cite{ESSPG_vs_ISSPG}. Both use the high-torque-density Vector Technics 8020 motor, with the optimization framework selecting the gearbox type and parameters.

The thigh and shank links adopt a sandwich-style structure\cite{Stoch3Design}, comprising a $3$D-printed plastic core enclosed between two laser-cut aluminum plates. While the aluminum layers provide structural strength, the plastic component defines the link’s geometry. This configuration enables low-cost fabrication with minimal precision machining and contributes to reduced inertia for improved dynamic performance.

\begin{figure}
    \centering    
    \includegraphics[width=0.7\linewidth]{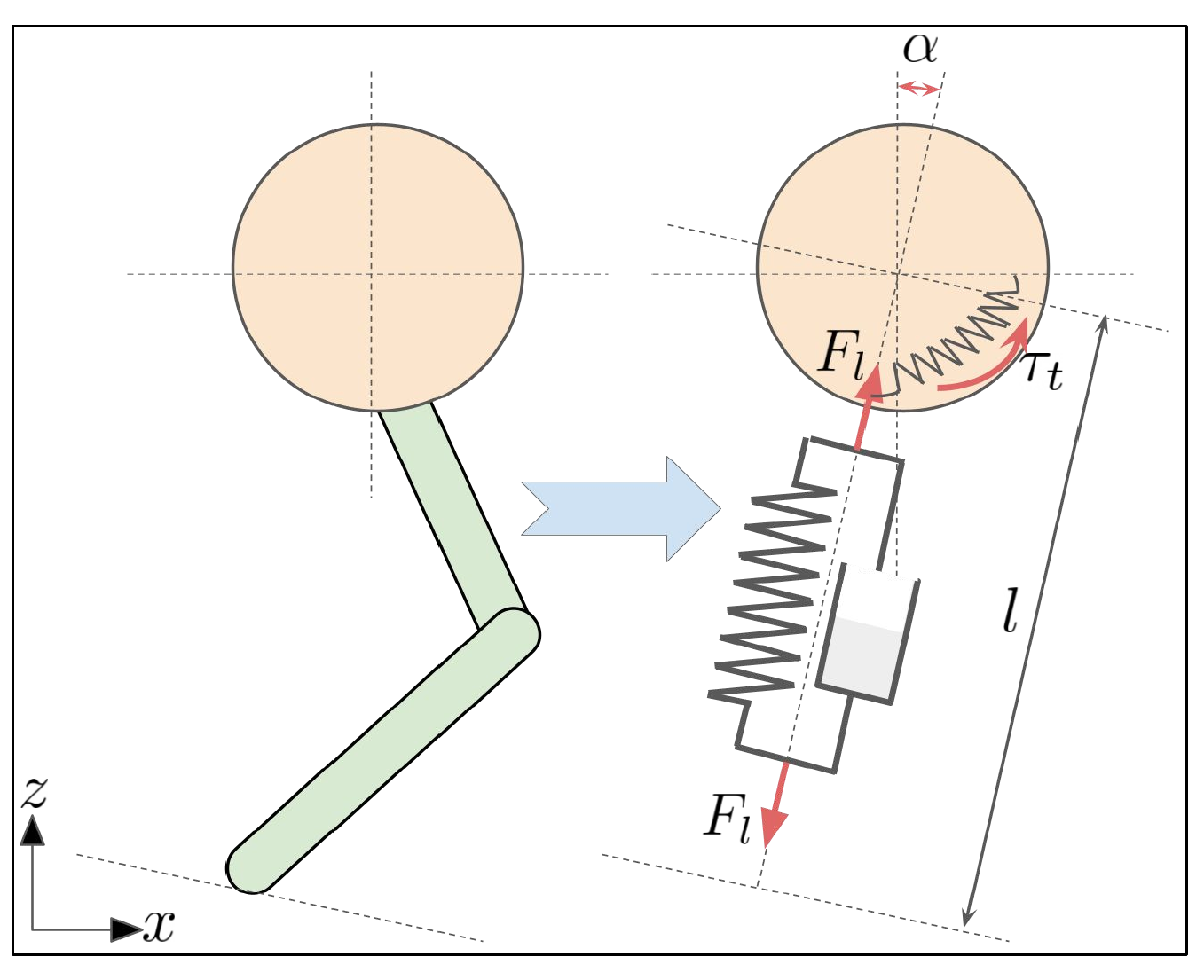}
    \caption{The system is modeled as a parallel linear spring–damper of length \(l\), the distance from the base center to the foot. A torsional spring models angular deflection about the world-frame y-axis from a fixed vertical (z-axis). The robot’s mass is assumed concentrated at the base.}
    \label{fig:sd}
    \vspace{-1em}
\end{figure}

\textbf{Leg link mass} is estimated using a parametric model that maps link length to mass. Based on the sandwich design, the model takes link length as input, computes the volume of aluminum plates, 3D-printed parts, chains, sprockets, and couplings, and multiplies each by its material density. The mass estimation process is shown in Fig.~\ref{fig:Mass_models}(b). This function outputs total link mass as a function of length, yielding estimates closely matching real values. The resulting mass-length relationship is nearly linear, as shown in Fig.~\ref{fig:mass_v_ll} .

% Each actuator follows a Quasi-Direct Drive (QDD) configuration, consisting of a brushless DC (BLDC) motor coupled with a single-stage planetary gearbox. Two actuator architectures are utilized: External Single-Stage Planetary Gearbox (ESSPG)~\cite{MitCheetah3} and Internal Single-Stage Planetary Gearbox (ISSPG)~\cite{MiniCheetah}. In the ESSPG architecture, the planetary gearbox is mounted outside the BLDC motor, whereas in the ISSPG, it is integrated within the motor stator. The ISSPG offers a lighter and more compact design but is limited in achievable gear ratios. In contrast, the heavier ESSPG can attain higher gear ratios. Both actuators use the Vector Techniques 8020 motor, selected for its high torque density. 
% An actuator optimization framework selects both the actuator type and its optimal gearbox parameters. 

% \subsubsection{Leg Link Mass Modeling}\label{sec:Leg_Link_Mass_Modeling}

\subsection{Monoped Control Architecture}
\label{Controller}

The jumping controller is based on a virtual spring–damper model adapted from \cite{VMC}, which models a symmetric 5-bar leg as a 1R–1P mechanism with a torsional spring–damper at the revolute joint and a linear spring–damper at the prismatic joint. In contrast, our serial 2R leg is modeled with only a torsional spring at the revolute joint and a linear spring–damper at the prismatic joint (Fig.~\ref{fig:sd}).   
% to emulate realistic interaction dynamics during ground contact. 

% \textcolor{red}{\textbf{Virtual Model Representation:}  
% The equivalent system comprises a linear spring and damper arranged in parallel along length $ l $, defined as the distance from the base center to the foot. A torsional spring accounts for angular deflection about the world-frame y-axis, calculated from a fixed vertical (z-axis, as shown in Fig. \ref{fig:sd}). The robot's mass is assumed to be concentrated at the base.}

% \textbf{Virtual Model Representation:}  
% The system is modeled as a parallel linear spring–damper of length \(l\), the distance from the base center to the foot. A torsional spring models angular deflection about the world-frame y-axis from a fixed vertical (z-axis, Fig.~\ref{fig:sd}). The robot’s mass is assumed concentrated at the base.

\textbf{Force and Torque Computation:}% The virtual spring has a resting length $ l_0 $, significantly greater than the initial value of $l$, ensuring it remains compressed during ground contact. The axial force generated by the spring-damper pair is given by, along length $l$:
% \begin{equation}
% \label{fl}
% F_{l}(l, \dot{l}) = K(l_{0} - l) - C\dot{l}
% \end{equation}
% where $K$ and $C$ are the linear spring and damping coefficients. The value of $l$ would depend on the angular positions of hip and knee joints, $\theta_{1}$ and $\theta_{2}$ respectively. The corresponding torque due to the torsional spring is:
% \begin{equation}
% \label{Tt}
% \tau_{t}(\alpha) = T(\alpha_{0} - \alpha)
% \end{equation}
% where $ \alpha $ is the current deflection and $ \alpha_0 $ is the resting orientation. Here, $T$ denotes the torsional spring constant. 
The virtual spring has a resting length $l_0$ greater than the initial $l$, ensuring compression during ground contact. The axial force from the spring-damper pair along $l$ is:  
\begin{equation}
\label{fl}
F_{l}(l, \dot{l}) = K(l_{0} - l) - C\dot{l}
\end{equation}
where $K$ and $C$ are the linear spring and damping coefficients. The length $l$ depends on hip and knee joint angles $\theta_{1}$ and $\theta_{2}$. The torque from the torsional spring is:  
\begin{equation}
\label{Tt}
\tau_{t}(\alpha) = T(\alpha_{0} - \alpha)
\end{equation}
where $\alpha$ is the deflection, $\alpha_0$ the resting orientation, and $T$ the torsional spring constant.

% In this case, $l_0$ and $\alpha_0$ used in equations \ref{fl} and \ref{Tt} are $10m$ and $0$ respectively.

\textbf{Force Decomposition in World Frame:}  
% The ground reaction forces in world-frame $z$ and $x$ directions are computed as:
The reaction forces exerted on the ground by the virtual spring-damper model are computed along the world frame $z$-axis and $x$-axis as follows:
\begin{equation}
\label{fz}
F_{z} = -F_{l}\cos\alpha + \frac{\tau_{t}}{l} \sin\alpha - mg
\end{equation}
\begin{equation}
\label{fx}
F_{x} = -F_{l}\sin\alpha - \frac{\tau_{t}}{l} \cos\alpha
\end{equation}

\textbf{Joint Torque Computation:}  
Joint torques are computed using $ \tau = J(\theta_1,\theta_2)^{T}F $, where $F = [F_{x}, F_{z}]^{T}$, and $J(\theta_1,\theta_2)$ is the Jacobian of the 2-DOF leg (base to hip). $\theta_{1}$ and $\theta_{2}$ are the hip and knee angular positions, respectively. These are obtained from the simulator. 

% \textcolor{red}{The controller parameters $K, C,$ and $T$ directly influence actuator torques and in turn, the maximum height in a jump and the mechanical energy consumed.}

% \textbf{Adaptive Parameter Tuning:}  
% The spring and damping coefficients are modulated based on the jumping phase. Higher spring and lower damping values are used during take-off; the reverse is applied during landing. The base velocity direction is used to determine the phase.

% The controller parameters $K, C,$ and $T$ directly influence actuator torques and jumping height. 

%The CMA-ES algorithm tunes these for each candidate design during grid search, as detailed in Section~\ref{Optimisation_framework}.

\section{Optimization Framework Overview} \label{opt_framework}
% This paper presents a three-stage framework for the co-optimization of design and control parameters of a monoped robot to achieve maximum jumping height with minimal energy consumption. The outcome includes both optimal parameters and a template CAD model suitable for fabrication.

This section presents a three-stage framework to co-optimize monoped design and control for maximum jump height and minimal energy use, producing optimal parameters and a template CAD model.

\begin{figure*}[htbp]
    \centering
    \includegraphics[width=0.95\textwidth]{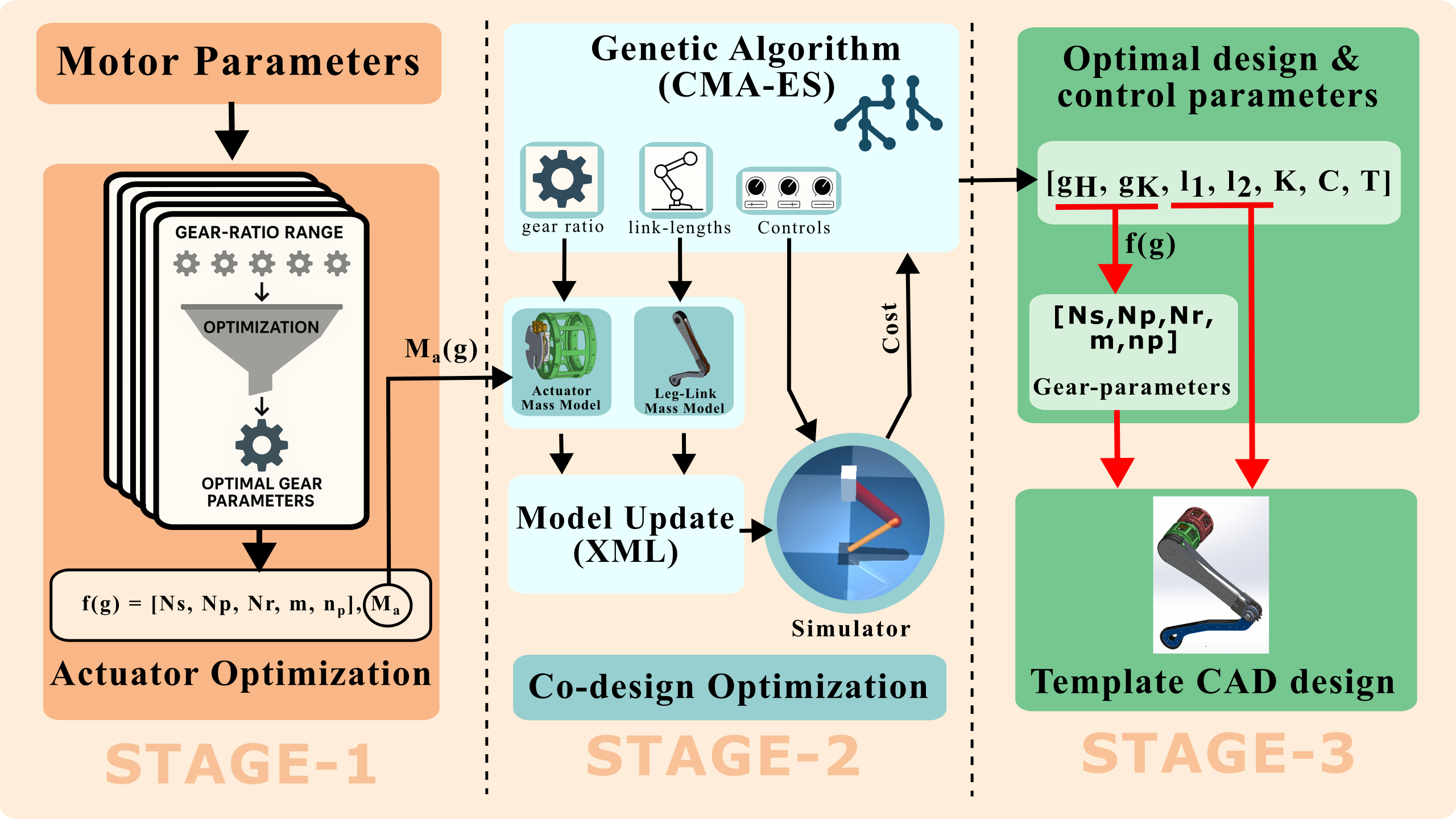} 
    \caption{Overview of the three-stage co-design framework. \textbf{Stage-1: Actuator Optimization} computes optimal gear parameters for a given motor, mapping gear ratio to actuator mass. \textbf{Stage-2: Co-design Optimization} uses CMA-ES to optimize gear ratios, link lengths, and control parameters. \textbf{Stage-3: Template CAD Generation} auto-generates a parametric CAD model for visualization and prototyping.}
    \label{fig:flowchart2}
    \vspace{-1em}
\end{figure*}

\subsection{Stage 1: Actuator-Level Optimisation}
This stage establishes a one-to-one mapping between gear ratio and the corresponding actuator mass and peak torque. As multiple designs with different gearbox architectures (ESSPG or ISSPG) and masses can yield the same ratio, the mapping is non-trivial.
% To resolve this, we formulate and solve an optimization problem over discrete gear ratio intervals (e.g., $4.0$:$1$-$4.1$:$1$, $4.1$:$1$-$4.2$:$1$, ..., $14.9$:$1$-$15.0$:$1$). For each interval, the framework identifies the lightest actuator design capable of achieving the target ratio, accounting for both gearbox type and design parameters. The result is a gear ratio-to-actuator mapping that selects the optimal actuator (in terms of minimal mass) across both ESSPG and ISSPG configurations.
We address this by solving an optimization problem over discrete gear ratio intervals (e.g., $4.0$:$1$–$4.1$:$1$, …, $14.9$:$1$–$15.0$:$1$). For each interval, the framework selects the lightest actuator meeting the target ratio, considering gearbox type and design parameters. This yields a gear ratio–actuator mapping that minimizes mass across ESSPG and ISSPG configurations.
% The actuator optimization methodology is shown in Fig.~\ref{fig:flowchart2}. This mapping is subsequently used in Stage 2 for jointly optimizing design and control parameters. The formulation of the optimization problem for obtaining the lightest actuator in a given ratio range is presented below.
The actuator optimization methodology is shown in Fig.~\ref{fig:flowchart2}. This mapping is then used in Stage 2 to jointly optimize design and control parameters. The optimization formulation for selecting the lightest actuator within a ratio range is presented below.

\subsubsection{Design Variables}
% For both ESSPG and ISSPG—being single-stage planetary gearboxes—the design variables are identical and can be represented by a column vector, $X \in \mathcal{X} \subseteq \mathbb{R}^5$, as follows:
% \begin{equation}\label{Opt_var}
%     X := [ N_{s}, N_{p}, N_{r}, m, n_p]
% \end{equation}
% where $N_{s}, N_{p}, N_{r}$ are the number of teeth on the sun, planet, and ring gears; $m$ is the gear module; $n_p$ is the number of planet gears; and $\mathcal{X}$ represents the vector space of design variables for this optimization stage. All variables except $m$ are integers, and $m$ takes values from a discrete set.

For both ESSPG and ISSPG single-stage planetary gearboxes, the design variables are identical and form a column vector \(X \in \mathcal{X} \subseteq \mathbb{R}^5\):
\begin{equation}\label{Opt_var}
    X := [ N_{s}, N_{p}, N_{r}, m, n_p]
\end{equation}
Here, \(N_{s}, N_{p}, N_{r}\) are the teeth counts of the sun, planet, and ring gears; \(m\) is the gear module; \(n_p\) is the number of planet gears; and \(\mathcal{X}\) is the design-variable space for this stage. All variables except \(m\) are integers, while \(m\) takes values from a discrete set.

\subsubsection{Constraints}
The optimization is subject to several constraints to ensure geometric compatibility, functional correctness, and manufacturability. These constraints are formally specified as follows:

\begin{enumerate}[label=\Roman*.]
% \item \textbf{Gear Ratio Constraint:} The gear ratio of single stage planetary gearbox can be given as, $GR = (N_{s} + N_r)/N_{s}$~\cite{Theory_of_Machines_and_Mechanism_RS_Khurmi}. This constraint ensures the gear ratio lies within a specified range for each interval. The constraint is mathematically represented as:
% \begin{equation}\label{eq:gear_ratio_constr}
%     GR_{min} \leq \frac{N_{s} + N_{r}}{N_{s}} \leq GR_{max}
% \end{equation}

\item \textbf{Gear Ratio Constraint:} The gear ratio of a single-stage planetary gearbox is \(GR = (N_{s} + N_{r})/N_{s}\)~\cite{Theory_of_Machines_and_Mechanism_RS_Khurmi}. 
The following constraint ensures it remains within the specified range:
\begin{equation}\label{eq:gear_ratio_constr}
    GR_{min} \leq \frac{N_{s} + N_{r}}{N_{s}} \leq GR_{max}
\end{equation}

\item \textbf{Geometric Constraint:} This constraint ensure the dimensional compatibility of the sun, ring, and, planet gears. It is mathematically represented~\cite{sspgTeethMatchingCond} as:
% \textcolor{blue}{Add a line about its implication. Use the same structure as used in the other constraints.}
\begin{equation}\label{eq:geometric_constr}
    N_r = N_s + 2N_p
\end{equation}

\item  \textbf{Meshing Constraint:} This ensures proper tooth engagement between the sun and ring gears. Its mathematical representation~\cite{sspgTeethMatchingCond} is:
\begin{equation}\label{eq:meshing_constr}
    N_s + N_r = n_p a,\quad a \in \mathbb{Z}^{+}
\end{equation}

\item  \textbf{No Interference Constraint:} This constraint is imposed to prevent collision between adjacent planet gears. Its mathematical representation~\cite{sspgTeethMatchingCond} is:
\begin{equation}\label{eq:no_interference_constr}
    2m (N_s + N_p) \sin\left(\frac{\pi}{n_p}\right) - 2mN_p \geq \delta_{p}
\end{equation}
where, $\delta_p = 5$ mm, in this work. 

\item  \textbf{Additional Constraints:} These constraints limit the range of the optimization variables. They are mathematically represented as:
% \textcolor{blue}{Add a line about its implication. Use the same structure as used in the other constraints.}
\begin{equation}\label{eq:additional_constr}
\begin{split}
    & m_{\min} \leq m \leq m_{\max},\quad N_s, N_p \geq N_{\min} \\
    & m N_r \leq D^{\max}_{GB},\quad n^{\min}_p \leq n_p \leq n^{\max}_p
\end{split}
\end{equation}
\end{enumerate}

We use $m_{\min} = 0.5$ mm, $m_{\max} = 1.2$ mm, and $N_{\min} = 18$ to avoid undercutting. The gearbox outer diameter is constrained as $D^{\max}_{GB} = D^{\text{motor}}_{OD} - \delta_{\text{clr}}$ for ESSPG, and $D^{\max}_{GB} = D^{\text{motor}}_{ID} - \delta_{\text{clr}}$ for ISSPG, with $\delta_{\text{clr}} = 10$ mm. The number of planet gears is limited to $n_p \in [2, 7]$ based on standard practices.

\subsubsection{Optimization Formulation}  

% The objective is to minimize actuator mass, $M_{act}$, for a given motor and gear ratio range. $M_{act} $ is computed using a detailed mass model that takes motor dimensions, motor mass, and gear parameters ($X$ ) as inputs. Based on these inputs, the dimensions of other actuator components—bearings, outer covering, carrier, coupling, backplate, etc.—are derived. Their volumes are then calculated and multiplied by respective material densities to obtain individual masses. All components, except the motor, bearings, and gears, are assumed to be made of aluminum. Bearing mass is estimated using a regression-based mass function derived from manufacturer datasheets, which takes the bearing’s internal diameter as input. The actuator mass model is depicted in Fig.~\ref{fig:Mass_models}(a).

The objective is to minimize actuator mass \(M_{act}\) for a given motor and gear ratio range. A higher actuator mass increases the required Ground Reaction Force, thereby raising torque demands. Minimizing actuator mass while achieving the required torque is critical for highly dynamic robots~\cite{MITActuatorDesign}. Accordingly, the optimization seeks the lightest actuator satisfying the motor and gear ratio constraints. $M_{act}$ is computed using a detailed model that inputs motor dimensions, motor mass, and gear parameters ($X$). From these, dimensions of other components, bearings, casing, carrier, coupling, backplate, etc., are derived. Volumes are calculated and multiplied by material densities to obtain masses. All parts except the motor, bearings, and gears are assumed aluminum. Bearing mass is estimated via a regression-based function from manufacturer datasheets using the bearing’s internal diameter. The actuator mass model is shown in Fig.~\ref{fig:Mass_models}(a).
Based on this objective, we formulate the final optimization problem as follows:
\begin{equation}
\begin{aligned}
    % \underset{X}{\text{Minimize:}} & \quad Cost \ Function \ (\ref{eq:cost}) \\
    \min_{\mathcal{X}} \quad & M_{\text{act}} \\
     \text{s.t.} \quad 
     & \text{Eqs.} ~\eqref{eq:gear_ratio_constr},~\eqref{eq:geometric_constr},~\eqref{eq:meshing_constr},~\eqref{eq:no_interference_constr},~\eqref{eq:additional_constr}
\end{aligned}
\end{equation}

% The solution $M_{act}^*$ to this optimization problem yields the minimum actuator mass for a given range of gear ratios. The problem is solved over gear ratio intervals from 4.0:1 to 15.0:1 in 0.1 increments for both ISSPG and ESSPG gearbox types, selecting the lightest actuator design across both types for each ratio interval. This produces a one-to-one mapping from gear ratio to actuator mass, which is used in Stage-2 of the monoped co-design optimization. A brute-force search method is employed, and the results are reported in Section~\ref{sec:ActuatorOptResults}. Peak torque of the actuator is calculated as the product of the gear ratio and the motor’s peak torque.

The solution $M_{act}^*$ gives the minimum actuator mass for a gear ratio range. The problem is solved over ratios from 4.0:1 to 15.0:1 in 0.1 increments for both ISSPG and ESSPG types, selecting the lightest design for each interval. This yields a gear ratio–mass mapping used in Stage-2 of the monoped co-design. A brute-force search is employed, and results are in Section~\ref{sec:ActuatorOptResults}. Peak actuator torque equals the gear ratio multiplied by the motor’s peak torque.

\subsection{Stage 2: Co-Design Optimization}
\label{co-opt problem} 
% The one-to-one mapping of gear ratio to mass, generated from Stage~1 of the framework is now utilized in Stage~2 of the proposed framework. In this stage, a co-design optimization of both mechanical design and control parameters is performed. This stage aims to maximize jump height while accounting for energy consumption. The co-design optimization methodology is shown in Stage 2 of Fig.~\ref{fig:flowchart2}. The optimization problem in this stage is formulated as follows: 

The gear ratio–mass mapping from Stage1 is used in Stage2 for co-design optimization of mechanical and control parameters. This stage seeks to maximize jump height while considering energy consumption. The methodology is shown in Stage2 of Fig.\ref{fig:flowchart2}, and the optimization problem is formulated as follows:

\subsubsection{Optimization variables}
\label{sec:co-opt var}
% The variables of the co-design optimization problem can be represented as a column vector, $Y \in \mathcal{Y} \subseteq{R^7}$, as follows:  
% \begin{equation}\label{co-opt_var}
%     Y := [ l_{1}, l_{2}, g_{k}, g_{h}, K, C, T]
% \end{equation}
% where $l_{1}, l_{2}$ are the thigh and shank link lengths respectively; $g_{h}, g_{k}$ are the gear ratios of hip and knee actuators respectively; $K, C, T$ are the control parameters (as explained in Section ~\ref{Controller}) and  $\mathcal{Y}$ is the vector space of design variables for this optimization stage. 

The co-design optimization variables form a column vector \(Y \in \mathcal{Y} \subseteq \mathbb{R}^7\):
\begin{equation}\label{co-opt_var}
    Y := [l_{1}, l_{2}, g_{k}, g_{h}, K, C, T]
\end{equation}
Here, \(l_{1}, l_{2}\) are thigh and shank lengths; \(g_{h}, g_{k}\) are hip and knee gear ratios; \(K, C, T\) are control parameters (see Section~\ref{Controller}); and \(\mathcal{Y}\) is the design-variable space for this stage.

\subsubsection{Constraints}
\label{sec:co-opt const}
The constraints for this problem are simply the bounds for all the variables:
\begin{equation}\label{eq:co-opt_constr}
\begin{split}
    & l_{\min} \leq l_{1}, l_{2} \leq l_{\max}, \quad
    g_{\min} \leq g_{h}, g_{k} \leq g_{\max},\\
    & K_{\min} \leq K \leq K_{\max}, \quad
    C_{\min} \leq C \leq C_{\max}, \\
    & \qquad \qquad \quad T_{\min} \leq T \leq T_{\max}    
\end{split}
\end{equation}
% We use $l_{\min} = 0.3$ m, $l_{\max} = 0.5$ m, $g_{\min} = 4$ and $g_{\max} = 8.7$. For control variables, the limits are $K_{\min} = 0$, $K_{\max} = 200$, $C_{\min} = 0$, $C_{\max} = 10$, $T_{\min} = 0$ and $T_{\max} = 20$. The lower limits for the control variables have been chosen to be $0$. The upper limits for $K$, $C$ and $T$ have been chosen based on heuristics. For $K$ and $T$, the upper limits were chosen as the maximum height was not changing by increasing the limits of these variables, due to torque limits of actuators. The upper limit of $C$ if increased beyond this value was impacting the maximum jumping height negatively. 

We use \(l_{\min} = 0.3\) m, \(l_{\max} = 0.5\) m, \(g_{\min} = 4\), and \(g_{\max} = 8.7\). Control variable limits are \(K_{\min} = 5\), \(K_{\max} = 200\); \(C_{\min} = 0\), \(C_{\max} = 10\); and \(T_{\min} = 0\), \(T_{\max} = 50\). Lower and upper bounds are heuristic. For \(K\) and \(T\), increasing limits beyond the upper bound values does not improve maximum height due to actuator torque limits; for \(C\), higher limits reduce maximum height.

\subsubsection{Optimization formulation} 
% The cost formulation for stage~2 is explained below: 
% \begin{equation}
% \label{comb cost}
%     C_{c}(\tau, \omega)= \lambda_{1}C_{h}(\tau) + \lambda_{2}{C_{e}(\tau, \omega)}
% \end{equation}
% Here, \(C_{e}\) is the energy consumed for $1$ jump. \(C_{h}\) is the jump height cost. $\lambda_{1}$ and $\lambda_{2}$ are weighing coefficients for each of the costs. The calculation of both \(C_{h}\) and \(C_{e}\) has been explained below. \(C_{h}\) is calculated as follows: 
The Stage~2 cost is defined as  
\begin{equation}
\label{comb cost}
    C_{c}(\tau, \omega) = \lambda_{1} C_{h}(\tau) + \lambda_{2} C_{e}(\tau, \omega)
\end{equation}
where \(C_{h}\) is the jump height cost, \(C_{e}\) is the energy consumed for one jump, and \(\lambda_{1}, \lambda_{2}\) are their respective weights. The calculations of \(C_{h}\) and \(C_{e}\) are detailed below:

% \begin{equation}
% \label{height cost}
%     C_{h}(\tau) = K_{h}e^{-h(\tau)}
% \end{equation}
% where $ h $ is the maximum height attained by the robot from the ground in a single jump. Height \(h\) is obtained from the simulator. \(K_{h}\) is a scaling constant. The value of this constant is chosen to ensure that the order of both \(C_{h}\) and \(C_{e}\) is similar in equation ~\ref{comb cost}. The variable \(\tau\) represents a matrix with columns of a sequence of torque inputs at the hip ($\tau_h \in \mathbb{R}^m$) and knee joints ($\tau_k \in \mathbb{R}^m$) over one jump of the robot. The values of these are obtained from the controller.  Specifically, $\tau$ is represented as: $\tau = \begin{bmatrix} \tau_h & \tau_k \end{bmatrix} \in \mathbb{R}^{m \times 2}$, where \(m\) denotes the number of simulation timesteps for a single jump of the monoped. Note that \(h$ is dependent on only \(\tau\) applied up to contact of the foot with the ground.  The control parameters \(K\), \(C\) and \(T\) directly influence \(\tau\).  

\begin{equation}
\label{height cost}
    C_{h}(\tau) = K_{h} e^{-h(\tau)}
\end{equation}
where \(h\) is the maximum jump height from the ground, obtained from the simulator, and \(K_{h}\) is a scaling constant chosen to keep \(C_{h}\) and \(C_{e}\) of similar order in~\eqref{comb cost}. The variable \(\tau = \begin{bmatrix} \tau_h & \tau_k \end{bmatrix} \in \mathbb{R}^{m \times 2}\) contains torque sequences for the hip (\(\tau_h \in \mathbb{R}^m\)) and knee (\(\tau_k \in \mathbb{R}^m\)) joints over one jump, with \(m\) as the number of simulation timesteps. \(h\) depends only on \(\tau\) applied until foot-ground contact. Control parameters \(K\), \(C\), and \(T\) directly influence \(\tau\). Equation~\ref{height cost} is used in place of the maximum height value to improve optimization accuracy and ensure reasonable convergence time.

The mechanical energy consumed during a jump is computed by summing hip and knee actuator energies over all timesteps. For actuator \(j \in \{h,k\}\) at timestep \(i\),
\begin{equation}
E_{j}(\tau_{j}(i),\omega_{j}(i)) =
\begin{cases}
\tau_j(i) \, \omega_j(i) \, dt, & \tau_j(i) \, \omega_j(i) > 0, \\
0, & \text{otherwise}.
\end{cases}
\end{equation}
This excludes regenerative energy, counting only motor power applied to the system. The total energy is
\begin{equation}
C_e(\tau,\omega) = \sum_{i=1}^{m} \big( E_h(i) + E_k(i) \big),
\end{equation}
where \(\tau_h, \tau_k\) and \(\omega_h, \omega_k\) are torques and angular velocities at the hip and knee, obtained from the simulator. In matrix form,
\[
\tau = \begin{bmatrix} \tau_h & \tau_k \end{bmatrix}, \quad
\omega = \begin{bmatrix} \omega_h & \omega_k \end{bmatrix} \in \mathbb{R}^{m \times 2}.
\]
The Stage~2 optimization problem is:
\begin{equation}
\begin{aligned}
    \min_{\mathcal{Y}} \quad & \lambda_{1}C_{h}(\tau) + \lambda_{2}{C_{e}(\tau, \omega)} \\
    \text{s.t.} \quad & \text{Eq.}~\eqref{eq:co-opt_constr}
\end{aligned}
\end{equation}

We solve the optimization problem using the Covariance Matrix Adaptation Evolution Strategy (CMA-ES)~\cite{CMAES}, which samples \(N\) candidate solutions from a multivariate normal distribution with an initial mean and step-size \(\sigma\) controlling the search radius. Each sample defines design and control parameters, with design variables generating XML files for the MuJoCo simulator~\cite{mujoco}.

Gear ratios update actuator mass and torque limits via the Stage~1 mass model, while link lengths update monoped link dimensions and masses using the leg-link mass model (Section~\ref{monoped arch}, Fig.~\ref{fig:Mass_models}(b)). The simulator evaluates each sample on a single-jump task, computes a cost, and CMA-ES updates the distribution mean \(\mu\) and step-size \(\sigma\) based on current-generation costs. 
% This repeats until convergence or a maximum of \(M=200\) iterations. 
% The initial mean vector is \(X_i = [0.4,\, 0.4,\, 6.0,\, 6.0,\, 50,\, 2.5,\, 10]\), step-size \(\sigma_i = 0.1\), and population size \(N = 8\).

\subsection{Stage 3: Template CAD Model Generation}
% Stage~2 determines the optimal design and control parameters for the monoped by balancing jump height and energy consumption. The design variables include thigh and shank link lengths and the gear ratios for the hip and knee actuators. These gear ratios are mapped to actuator parameters—gear tooth count, module, number of planet gears, and facewidth—using the mappings from Stage~1. Template CAD model generation based on these parameters is shown in Fig.~\ref{fig:flowchart2}.

% Using the optimal actuator parameters and link lengths from Stage2, Stage3 automatically generates a parameterized CAD model of the monoped, including leg and actuator subassemblies. The leg follows a sandwich-style design architecture, as described in Section \ref{monoped arch}, where the geometric dimensions of each part are parameterized as functions of the respective link lengths. Similarly, actuator components such as bearings, carriers, backplates, and outer casings are dimensioned based on the derived gear design parameters.

% This automated framework generates a complete template CAD model of the monoped, enabling rapid visualization, design verification, and serving as a foundation for detailed mechanical development. 
% By automating key design decisions, it significantly accelerates the CAD process and minimizes manual iteration. 

Using optimal actuator parameters and link lengths from Stage 2, Stage 3 generates a parameterized CAD model of the monoped, including leg and actuator subassemblies. The leg employs a sandwich-style design \ref{monoped arch}, with part dimensions defined by link lengths, while actuator components such as bearings, carriers, backplates, and casings are dimensioned from gear parameters. This automated framework delivers a complete CAD template for rapid visualization, design verification, and as a foundation for detailed development, significantly accelerating the process and reducing manual iteration.

% \vspace{-1em}
\section{Results and Discussion}
\label{results}
% All experiments were performed on a machine with an 11th Gen Intel Core i9-11900K @ 3.50GHz (16 cores) CPU and 128GB RAM. %Each experiment (case) required approximately 30 minutes to complete.
%%%%%%%%%%%%%%%%%%%%%%%%%%%%%%%%%%%%%%%%%%%%%%%%%%%%%%%%%%%%%%%%%%%%%%%%%%

\begin{figure}[h]
    \centering
    \includegraphics[width=0.45\textwidth]{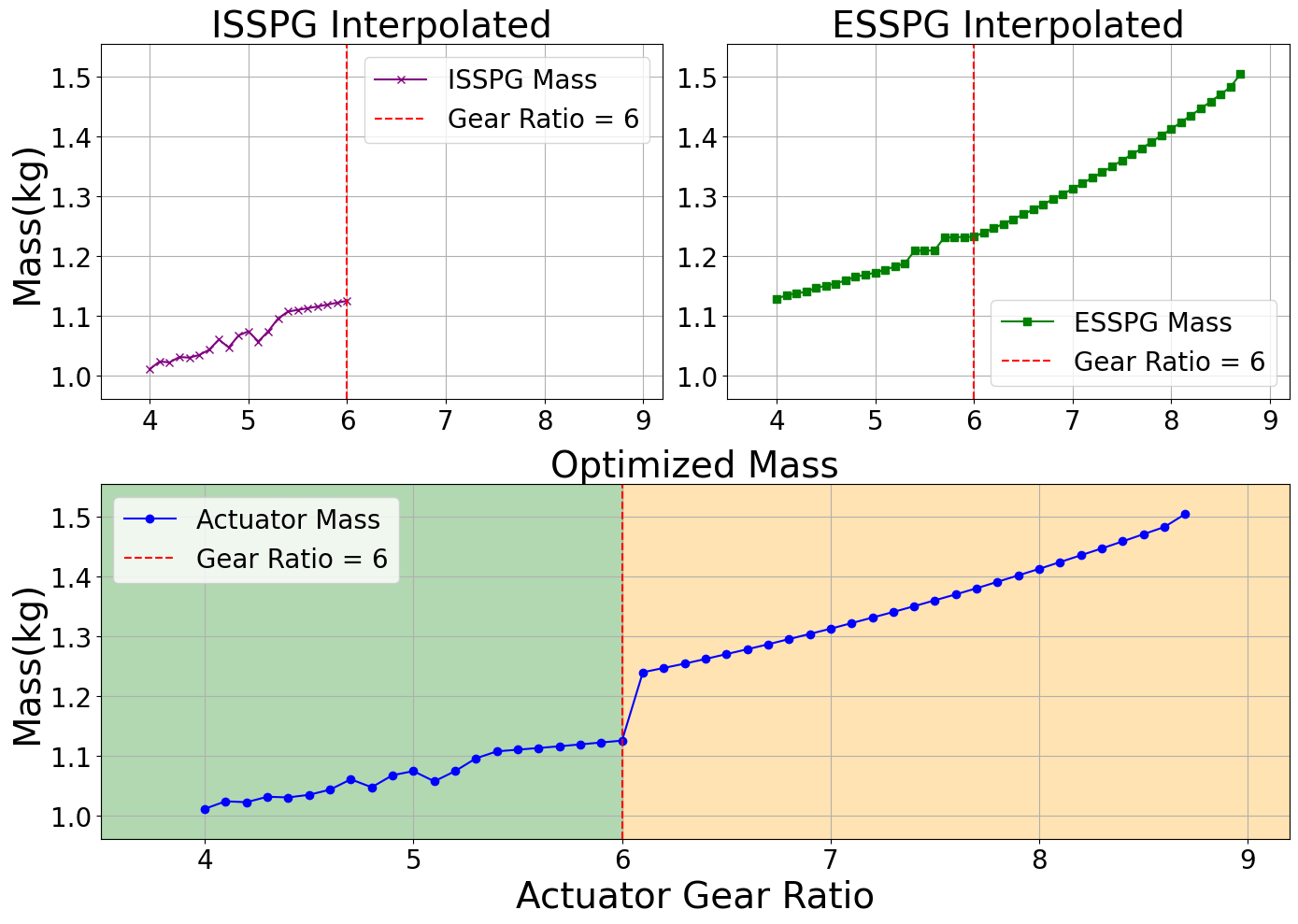}
    \caption{Mapping of gear-ratio to mass of the actuator. This is a result of the Stage-1 Optimization.}
    \label{fig:GR_v_Mass}
    \vspace{-1em}
\end{figure}

\subsection{Actuator Optimization Results}\label{sec:ActuatorOptResults}

% This subsection presents the results of Stage-1 of the optimization framework, which solves the actuator optimization problem across various gear-ratio ranges to establish a one-to-one mapping between gear ratio and gearbox parameters such as number of teeth, module, and number of planets. Actuator mass includes contributions from the motor, gears, couplings, carriers, outer casing, backplate, and bearings. Since these dimensions depend on motor size and gearbox parameters, actuator mass becomes a function of gear ratio and motor properties. The optimization is performed for both ISSPG and ESSPG types. For ISSPG, gear ratios up to 6:1 are feasible due to constraints on the ring gear’s outer diameter, which must fit within the motor stator. ESSPG actuators, with external gearboxes, support ratios up to 8.7:1, limited by the motor’s outer diameter. ESSPG is heavier than ISSPG below 6:1 due to architectural differences. Thus, the combined mass model selects ISSPG for gear ratios below 6:1 and ESSPG above 6:1, as shown in Fig.~\ref{fig:GR_v_Mass}.

This subsection presents Stage-1 results, where the actuator optimization problem is solved across gear-ratio ranges to map gear ratio to gearbox parameters such as teeth count, module, and number of planets. Actuator mass includes the motor, gears, couplings, carriers, casing, backplate, and bearings, with dimensions determined by motor size and gearbox parameters, making mass a function of gear ratio and motor properties. Optimization is performed for ISSPG and ESSPG types. ISSPG ratios are limited to 6:1 by the ring gear’s outer diameter fitting within the stator, while ESSPG ratios reach 8.7:1, constrained by the motor’s outer diameter. ESSPG is heavier below 6:1 due to its architecture; thus, the model selects ISSPG for ratios under 6:1 and ESSPG above, as shown in Fig.~\ref{fig:GR_v_Mass}.

\subsection{Co-Design Optimisation Results}
\label{co-opt results}

\begin{figure}
    \centering
    \includegraphics[width=0.8\linewidth]{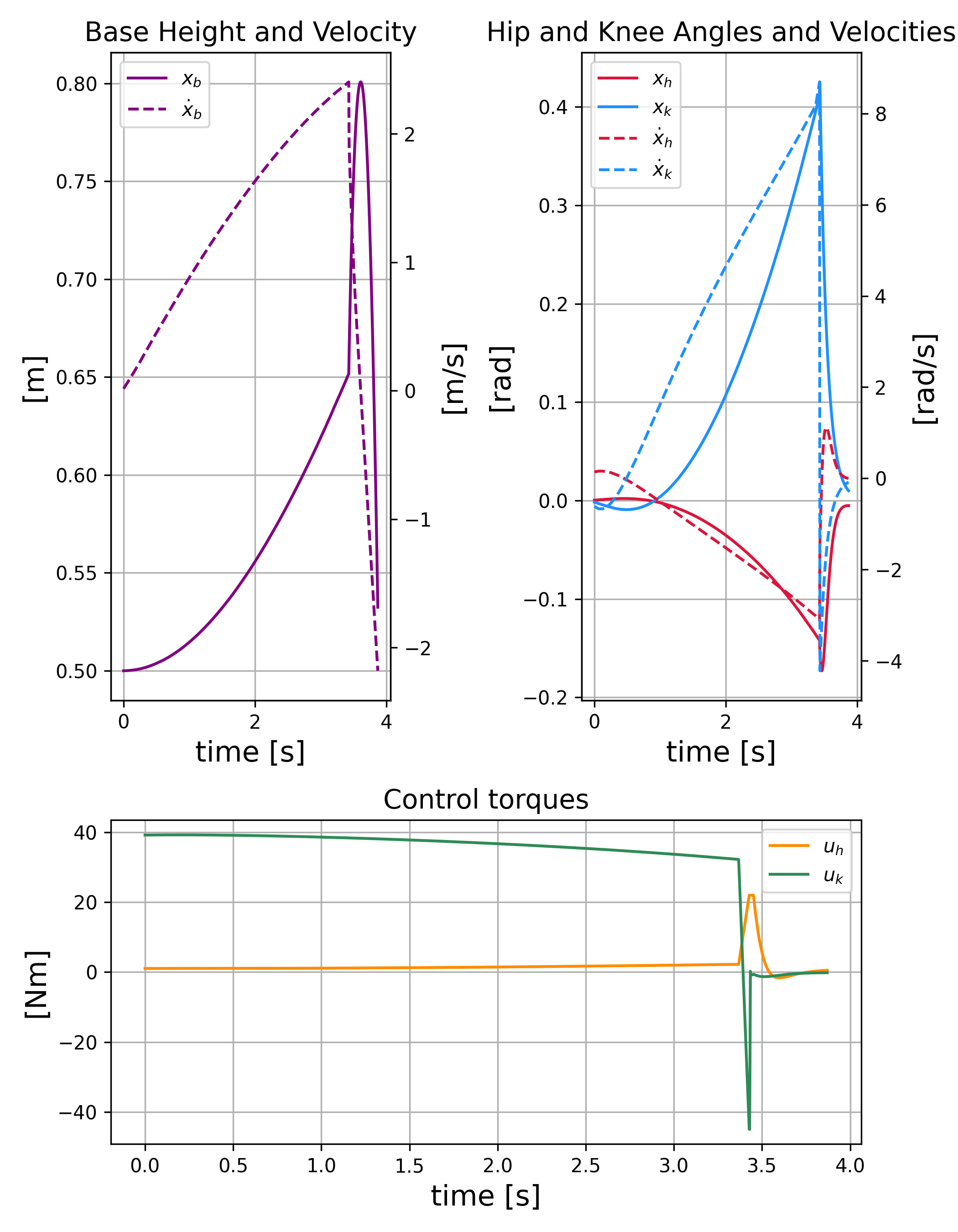}
    \caption{Variation of States \& control inputs with time for jump with optimized parameters, Case C}
    \label{fig:opt_parameters}
    \vspace{-1em}
\end{figure}

\begin{table*}[ht!]
\centering
\resizebox{0.8\textwidth}{!}{\begin{tabular}{|l|c|c|c|c|c|c|c|c|c|}
\hline
\textbf{Case} & \textbf{$\mathbf{g_{h}}$} & \textbf{$\mathbf{g_{k}}$} & \textbf{$\mathbf{l_{1}(m)}$} & \textbf{$\mathbf{l_{2}(m)}$} & \textbf{K} & \textbf{C} & \textbf{T} & \textbf{H(m)} & \textbf{E(J)} \\
\hline
Nominal & 6:1 & 6:1 & 0.4   & 0.4   & 50   & 2.5 & 10.0 & 0.887 & 31.44 \\
Case A       & 4:1 & 7.9:1 & 0.4   & 0.4   & 13.7 & 6.5 & 46.5 & 0.743 & 15.51 \\
Case B       & 6:1 & 6:1 & 0.499 & 0.346 & 8.9  & 1.0 & 6.9  & 0.771 & 17.02 \\
\textbf{Case C}       & \textbf{4:1} & \textbf{6:1} & \textbf{0.49} & \textbf{0.36} & \textbf{7.9}  & \textbf{2.2} & \textbf{8.4}  & \textbf{0.800} & \textbf{15.36} \\
\hline
\end{tabular}}
\caption{{\large Co-Design Optimization Results}}
\label{tab:Final_CoDesign_Results}
\vspace{-1em}
\end{table*}

% This subsection presents the optimized design and control variables obtained from the proposed framework. Jump height and mechanical energy consumption are first evaluated for a nominal monoped design during vertical jumping, and then compared across three optimization scenarios: Case-A, Case-B, and Case-C. In Case-A, hip and knee gear ratios and control parameters are optimized, with link lengths fixed to nominal values. Case-B optimizes thigh and shank link lengths along with control parameters, keeping gear ratios constant. Case-C jointly optimizes link lengths, gear ratios, and control parameters.These cases enable analysis of the influence of various design parameters on jump height and energy consumption. The gear ratio-to-actuator mass mapping used in these experiments is derived from the Actuator Optimization stage, as detailed in the previous subsection.

This subsection presents the optimized design and control variables from the proposed framework. Jump height and mechanical energy consumption are evaluated for a nominal monoped during vertical jumping and compared across three cases: Case-A, Case-B, and Case-C. Case-A optimizes hip and knee gear ratios with control parameters, keeping link lengths nominal. Case-B optimizes thigh and shank lengths with control parameters, keeping gear ratios fixed. Case-C jointly optimizes link lengths, gear ratios, and control parameters. These cases analyze the impact of design parameters on jump height and energy use. The gear ratio–mass mapping used here is from the Actuator Optimization stage detailed earlier.
% The nominal design and control parameters are $[g_h, g_k, l_1, l_2, K, C, T] = [6, 6, 0.4, 0.4, 50, 2.5, 10]$. The corresponding gear parameters for both hip and knee actuators are $[N_s, N_p, N_r, m, n_p] = [18, 36, 90, 0.5, 3]$, with ISSPG type.
The nominal design and control parameters are $[g_h, g_k, l_1, l_2, K, C, T] = [6, 6, 0.4, 0.4, 50, 2.5, 10]$. Both hip and knee actuators have gear parameters $[N_s, N_p, N_r, m, n_p] = [18, 36, 90, 0.5, 3]$ with an ISSPG type.

% Please note, for all designs sampled using the CMA-ES algorithm, the initial base height ($h_{0}$) is fixed at 0.5\,m to eliminate its effect on maximum jump height~\cite{PantherLeg}. The foot is initially positioned directly beneath the base to remove the influence of foot offset. Initial joint configurations are computed using inverse kinematics. The simulation timestep used in the MuJoCo simulator for the cost calculation defined in Equation~\ref{comb cost} is 0.002\,s. Controller torques are applied only when the robot is in contact with the ground to facilitate jumping.

For all designs sampled using the CMA-ES algorithm, the initial base height ($h_{0}$) is fixed at 0.5m to remove its effect on maximum jump height~\cite{PantherLeg}. The foot starts directly under the base to eliminate foot offset effects, and initial joint configurations are obtained via inverse kinematics. The MuJoCo simulation timestep for the cost in Equation~\ref{comb cost} is 0.002s. Controller torques are applied only when the robot contacts the ground to enable jumping.

\subsubsection{Case-A}

To assess gear ratio effects on jumping performance, link lengths were fixed while optimizing only gear ratios and control parameters. The optimal hip gear ratio, $\mathbf{g_h} = 4{:}1$, near the lower bound, reflects the hip’s low torque demand, reducing actuator mass and energy use with minimal impact on jump height. The optimal knee gear ratio, $\mathbf{g_k} = 7.9{:}1$ (ESSPG type), provides higher torque at the cost of added mass. As the knee supplies most jump torque, a higher ratio is advantageous, though it remains below the upper bound due to mass penalties. Table~\ref{tab:Final_CoDesign_Results} lists the optimized gear ratios and control parameters for Case-A: hip actuator—$[N_s, N_p, N_r, m, n_p] = [30, 30, 90, 0.5, 5]$ (ISSPG, $\mathbf{g_h} = 4{:}1$); knee actuator—$[N_s, N_p, N_r, m, n_p] = [22, 65, 152, 0.5, 3]$ (ESSPG, $\mathbf{g_k} = 7.9{:}1$). Optimizing gear ratios alone reduced energy consumption by $50.6\%$ but decreased maximum jump height by $16.2\%$.

\subsubsection{Case-B}
% In Case-B, we analyze the effect of link-length modification on maximum height attained and mechanical energy consumption by fixing the hip and knee gear ratios to their nominal values and optimizing only the link lengths and control parameters. The optimized design results in a longer thigh and shorter shank compared to the nominal configuration. This adjustment slightly reduces maximum height attained (from $0.887$\,m to $0.771$\,m) but significantly lowers energy consumption by nearly 45\% (from $31.44$\,J to $17.02$\,J). These results suggest that reconfiguring link lengths in this manner effectively reduces energy expenditure. The optimized values are presented in Table~\ref{tab:Final_CoDesign_Results}. As in this case there is no change in gear ratio the gear-parameters are same as the nominal design.

In Case-B, the effect of link-length modification on jump height and energy consumption is examined by fixing the hip and knee gear ratios to their nominal values and optimizing only link lengths and control parameters. The resulting design features a longer thigh and shorter shank compared to the nominal configuration, reducing energy consumption by $45\%$ (from $31.44$ J to $17.02$ J). However, this reconfiguration also decreases jump height by $13\%$ (from $0.887$ m to $0.771$ m). The optimized values are listed in Table~\ref{tab:Final_CoDesign_Results}. Since gear ratios remain fixed, the actuator parameters are identical to those in the nominal design.

\subsubsection{Case-C}

In Case-C, representing our proposed framework, actuator gear ratios, link lengths, and control parameters are jointly optimized. The hip gear ratio converges to the lower bound, $\mathbf{g_h} = 4{:}1$, due to its low torque demand, minimizing mass, while the knee reaches the ISSPG upper limit ($6{:}1$) to handle the primary jumping load without ESSPG’s added mass. Link lengths mirror Case-B’s trend of a longer thigh and shorter shank, likely improving energy efficiency. This setup reduces energy consumption by $51.1\%$ (from $31.44$ J to $15.36$ J) and achieves greater jump heights than Cases A and B, though $9.8\%$ below nominal (from $0.887$ m to $0.800$ m). The optimized parameters are in Table~\ref{tab:Final_CoDesign_Results}: hip—$[N_s, N_p, N_r, m, n_p] = [30, 30, 90, 0.5, 5]$ (ISSPG, $\mathbf{g_h} = 4{:}1$); knee—$[18, 36, 90, 0.5, 3]$ ($\mathbf{g_k} = 6{:}1$).

\begin{figure}[h]
    \centering    
    \includegraphics[width=0.35\textwidth]{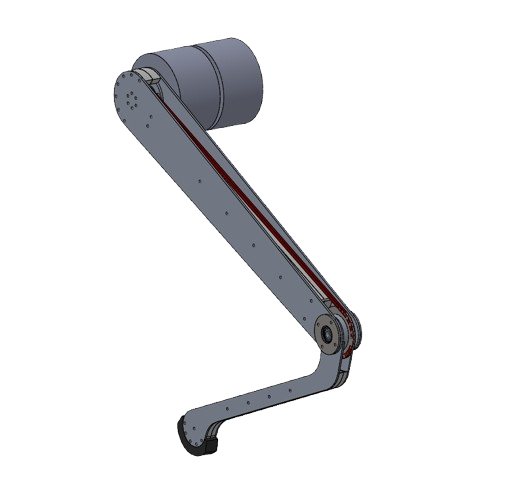}
    \caption{Template CAD model generated by the automated design generator for the Optimal monoped design}
    \label{fig:Template_CAD_model}
    \vspace{-1em}
\end{figure}

\subsection{Template CAD model generation}
% This section presents the template CAD model of the optimized monoped design. The model is parametric and generated automatically based on the optimal design variables obtained from the optimization framework. It takes the optimized link lengths and gearbox parameters to update the dimensions of the actuator components and the thigh and shank links, which in turn update the CAD model. This template aids in visualizing and verifying the monoped design and serves as a reference for human designers, enabling the creation of physical prototypes with minimal manual modifications. The optimization framework also reduces key design decisions, such as selecting the gear ratio, gearbox parameters, gearbox type, and link lengths. The resulting template CAD model is shown in Fig.~\ref{fig:Template_CAD_model}. This has been generated from results obtained in Case C of section \ref{co-opt results}. 

This section presents the parametric template CAD model of the optimized monoped, generated automatically from the design variables obtained in the optimization framework. The model updates actuator components and thigh–shank link dimensions using optimized link lengths and gearbox parameters. It facilitates visualization, verification, and rapid prototyping with minimal manual adjustments. The framework also finalizes key design choices, including gear ratio, gearbox parameters and type, and link lengths. The template CAD model, shown in Fig.\ref{fig:Template_CAD_model}, is generated from the Case-C results in Section \ref{co-opt results}.

% \vspace{-1em}
\section{Conclusion}
\label{conclusion}
% In this work, we presented a comprehensive three-stage co-design optimization framework for monopeds, targeting the simultaneous maximization of jump height while minimizing mechanical energy consumption. By integrating realistic actuator mass models and optimizing both mechanical (including gearbox, gear ratios, and link lengths) and control parameters, our approach bridges the gap between simulation-optimized designs and physical manufacturability. The automatic generation of parameterized CAD models further streamlines the transition from concept to hardware, significantly reducing manual design effort. Simulation results show the effectiveness of our framework, demonstrating effective co-optimization of maximum achievable jump height and energy consumption. Future work will extend this methodology to more complex multi-legged systems.

In this work, we proposed a three-stage co-design optimization framework for monopeds to simultaneously maximize jump height and minimize mechanical energy consumption. By incorporating realistic actuator mass models and jointly optimizing mechanical (gearbox, gear ratios, and link lengths) and control parameters, the framework bridges the gap between simulation and manufacturable designs. Automated generation of parameterized CAD models further streamlines the concept-to-hardware transition, reducing manual effort. Simulations confirm the framework’s effectiveness in co-optimizing jump height and energy use. Future work will extend the approach to multi-legged systems.
\bibliographystyle{IEEEtran}
\bibliography{references}

\end{document}